\documentclass[10pt,twocolumn,letterpaper]{article}

\usepackage[pagenumbers]{wacv} %

\usepackage{graphicx}
\usepackage{amsmath}
\usepackage{amssymb}
\usepackage{booktabs}
\usepackage{comment}

\usepackage[table,xcdraw]{xcolor}
\usepackage{times}
\usepackage{epsfig}
\usepackage{float}
\usepackage{multirow}
\usepackage{amsfonts}
\usepackage{graphics}
\usepackage{witharrows}
\usepackage{arydshln}

\usepackage[pagebackref,breaklinks,colorlinks]{hyperref}

\usepackage[capitalize]{cleveref}
\crefname{section}{Sec.}{Secs.}
\Crefname{section}{Section}{Sections}
\Crefname{table}{Table}{Tables}
\crefname{table}{Tab.}{Tabs.}

\newcommand{\fref}[1]{Figure \ref{#1}}
\newcommand{\tref}[1]{Table \ref{#1}}

\newcommand{\eref}[1]{Equation \ref{#1}}

\begin{document}

\title{HalluciDet: Hallucinating RGB Modality for Person Detection Through Privileged Information}

\author{Heitor Rapela Medeiros \\
{\tt\small heitor.rapela-medeiros.1@ens.etsmtl.ca}
\and
Fidel A. Guerrero Pe\~{n}a\\
{\tt\small fidel-alejandro.guerrero-pena@etsmtl.ca}
\and 
Masih Aminbeidokhti\\
{\tt\small masih.aminbeidokhti.1@ens.etsmtl.ca}
\and 
Thomas Dubail \\
{\tt\small thomas.dubail.1@ens.etsmtl.ca}
\and 
Eric Granger \\
{\tt\small eric.granger@etsmtl.ca}
\and 
Marco Pedersoli \\
{\tt\small marco.pedersoli@etsmtl.ca}
\\
\and
LIVIA, Dept. of Systems Engineering. ETS Montreal, Canada
}

\maketitle
\thispagestyle{empty}

\begin{abstract}
A powerful way to adapt a visual recognition model to a new domain is through image translation. However, common image translation approaches only focus on generating data from the same distribution as the target domain. Given a cross-modal application, such as pedestrian detection from aerial images,  with a considerable shift in data distribution between infrared (IR) to visible (RGB) images, a translation focused on generation might lead to poor performance as the loss focuses on irrelevant details for the task. In this paper, we propose HalluciDet, an IR-RGB image translation model for object detection. Instead of focusing on reconstructing the original image on the IR modality, it seeks to reduce the detection loss of an RGB detector, and therefore avoids the need to access RGB data. This model produces a new image representation that enhances objects of interest in the scene and greatly improves detection performance. We empirically compare our approach against state-of-the-art methods for image translation and for fine-tuning on IR, and show that our HalluciDet improves detection accuracy in most cases by exploiting the privileged information encoded in a pre-trained RGB detector. Code: \url{https://github.com/heitorrapela/HalluciDet}.
\end{abstract}

\begin{figure}[t]
\centering
    \resizebox{\columnwidth}{!}{%

    \begin{tabular}{cc}
    
    \toprule

    \includegraphics[width=.6\columnwidth]{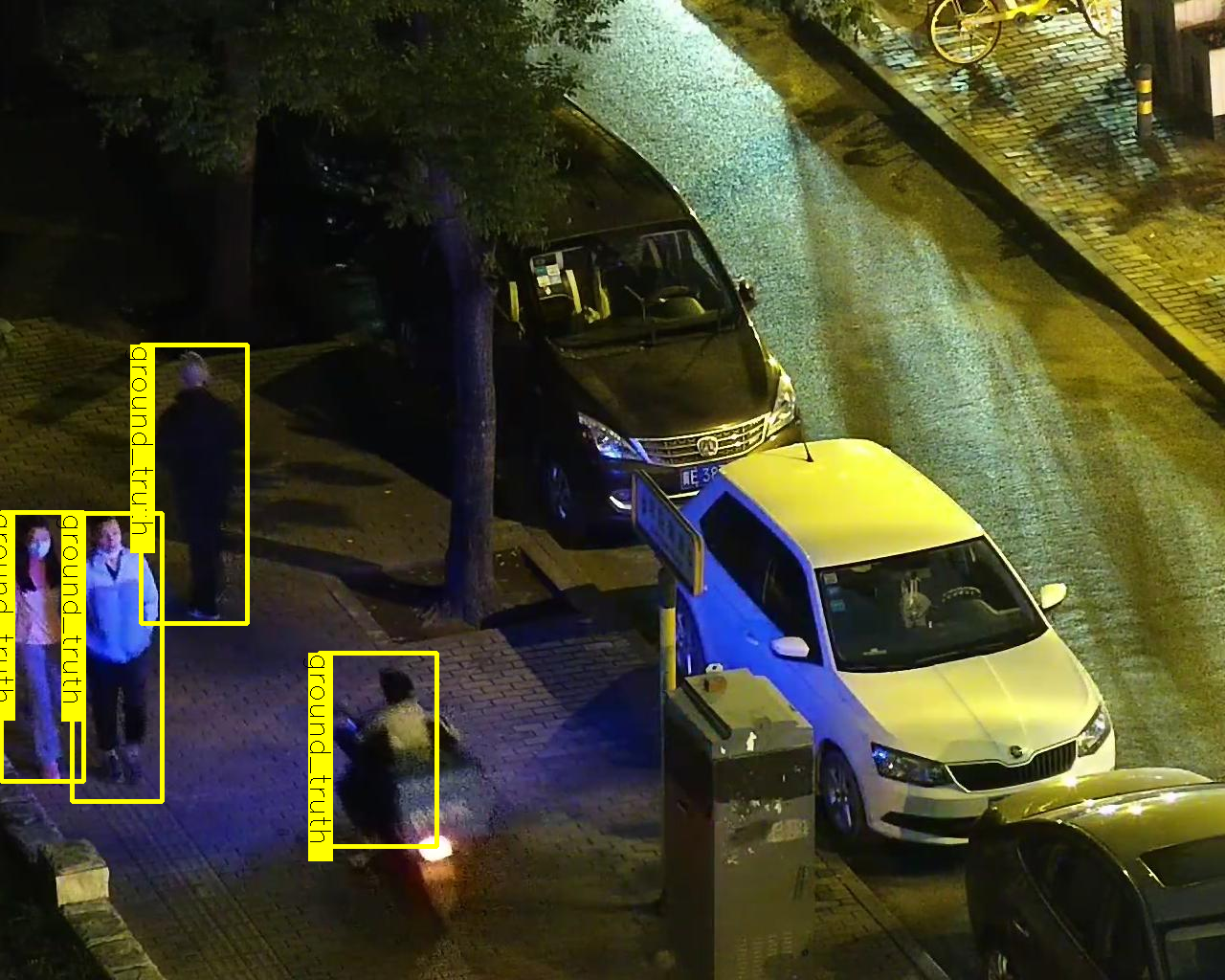} & 
    
    \includegraphics[width=.6\columnwidth]{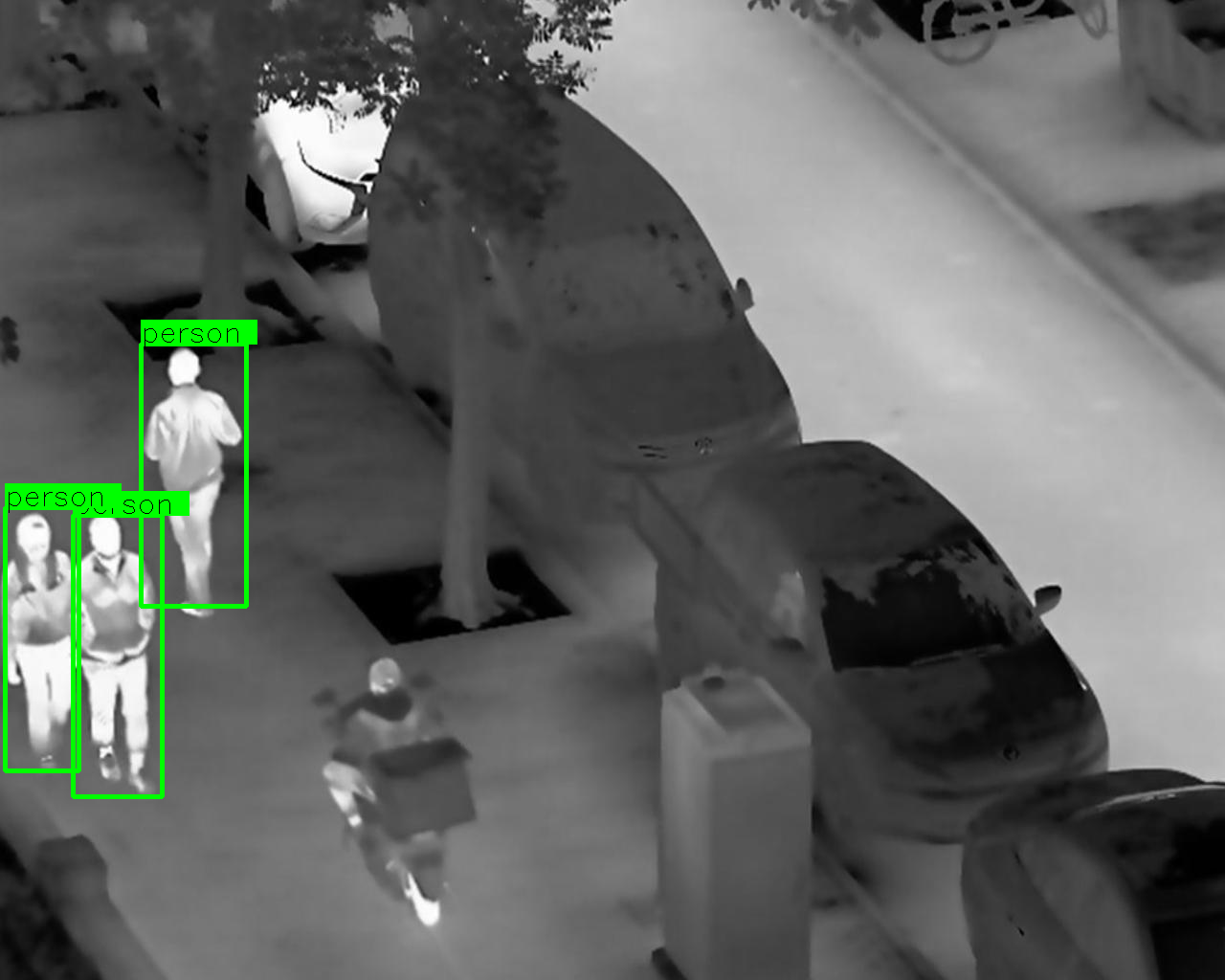} \\
    a) RGB image - Ground truth &  b) IR - Fine-tuned detections\\
  
    \includegraphics[width=.6\columnwidth]{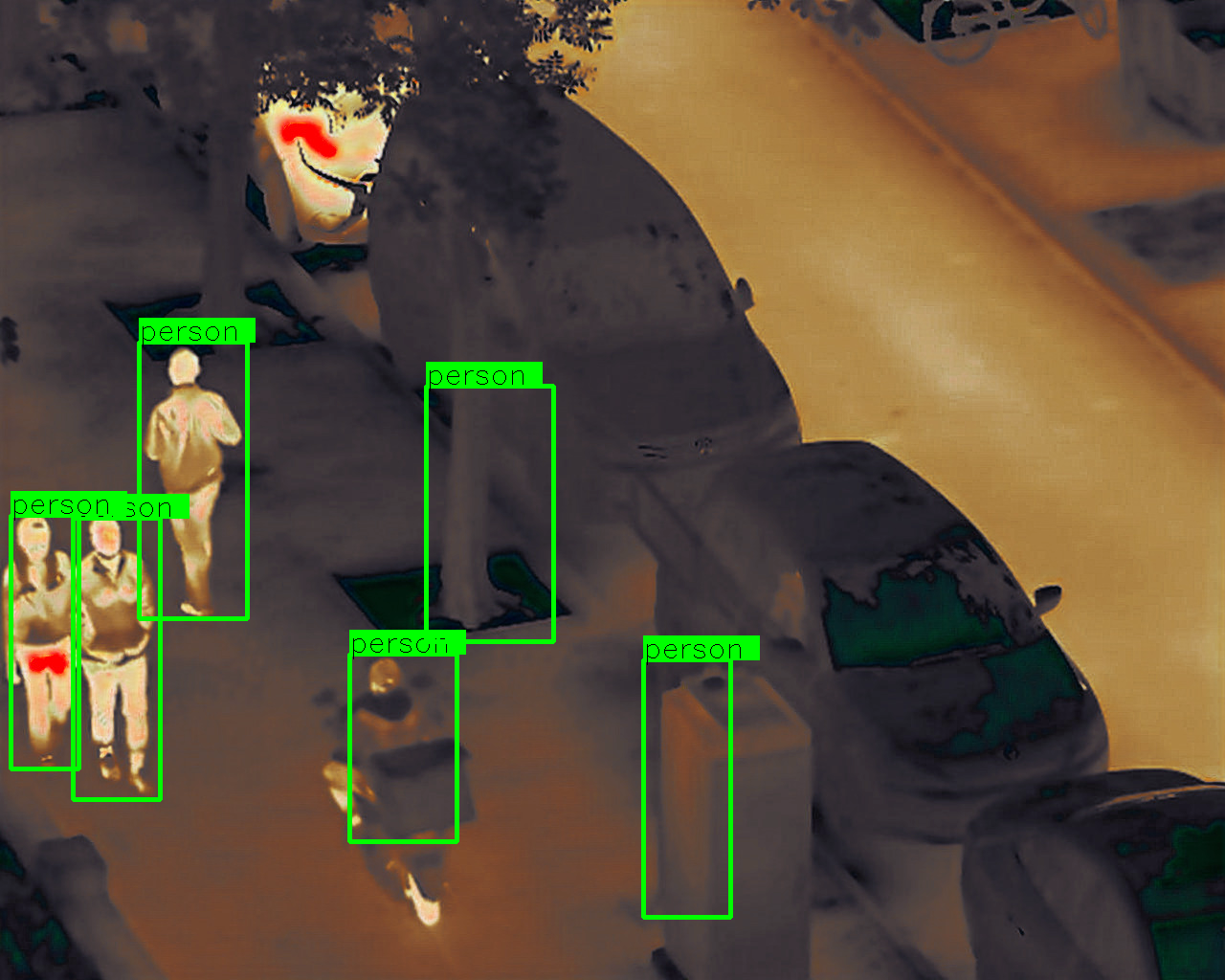} &

    \includegraphics[width=.6\columnwidth]{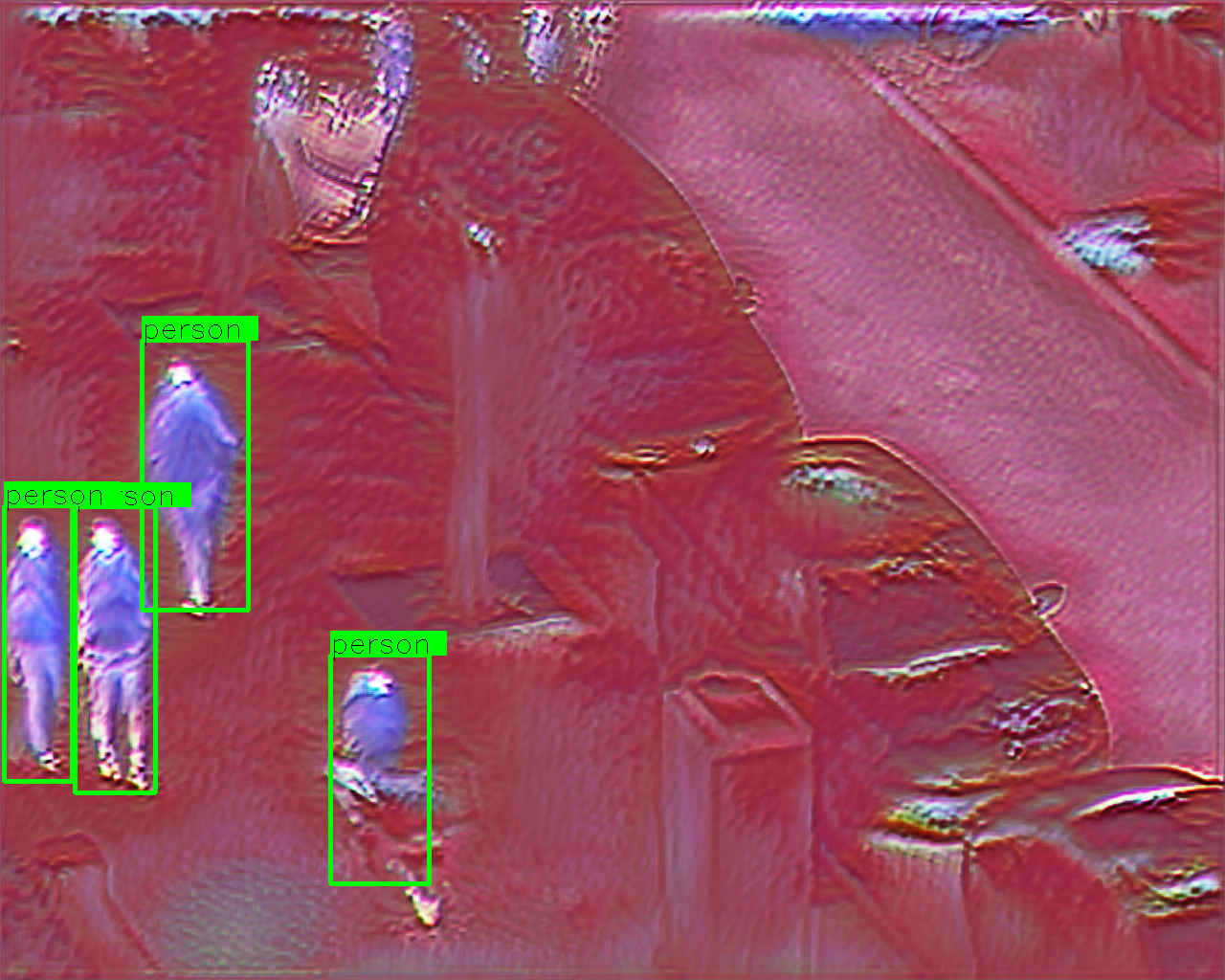} \\

    c) FastCUT - RGB detections & d) HalluciDet - RGB detections\\
    
    \bottomrule

    \end{tabular}
    }
\caption{Example of detections using baseline and HalluciDet methods on LLVIP data. (a) Original RGB image with ground truth annotations (yellow). (b) IR image with corresponding detections of a fine-tuned model (green). (c) Translated image from IR to RGB produced by FastCUT and corresponding RGB detections (green). (d) Hallucinated image produced by our method and RGB detections (green); HalluciDet does not seek to reconstruct all image details but only to enhance the objects of interest.}
\label{fig:04.04_qualitative_experiments123123}
\end{figure}

\section{Introduction} \label{sec:intro}

The proliferation of hardware sensors has greatly advanced the collection of large-scale datasets. Such datasets have significantly improved the performance of deep learning (DL) algorithms across various fields, including surveillance~\cite{chen2019distributed}, industrial monitoring~\cite{kong2021deep}, self-driving cars~\cite{stilgoe2018machine}, and robotics~\cite{pierson2017deep}. By providing high-resolution data, these sensors offer additional observations of common environmental phenomena to aid in the effectiveness of DL algorithms~\cite{ramachandram2017deep}. 

The additional information from different sensors has been employed in diverse settings~\cite{samaras2019deep, huang2020survey}. In computer vision applications, combining sensors with distinct environmental sensing perspectives, such as varying points of view and modality sensing information, can increase model performance, enabling possibilities that were previously unavailable. Furthermore, in the context of self-driving cars and intelligent building applications, two modalities commonly used are visible (RGB) and infrared (IR)~\cite{takumi2017multispectral}. In particular, the RGB modality offers valuable information for tasks like object detection, which generates bounding boxes for target objects within colored images. These colored images are known to have more diverse information due to their characteristics on the RGB light spectrum, especially in the presence of light. Thus, these RGB sensors are preferred to be used in daily activities where there is the presence of sunlight. On the other hand, the IR spectrum provides additional information for the visible modality when the light is low, especially during the night~\cite{jia2021llvip}, and also complementary information, primarily related to thermal sensing. Furthermore, IR is vastly applied in surveillance applications~\cite{zhang2018novel}, which require the device to capture information in light-restricted environments. IR object detection is known to detect objects using IR radiation emitted from the object, which varies depending on the object's material.

Despite the impressive performance of DL models, their effectiveness can significantly deteriorate when applied to modalities that were not present during the training~\cite{dai2018dark, torralba2011unbiased}. For example, a model trained on RGB images may not perform well on IR images during testing~\cite{yang2020reducing}. To address the issue, some studies utilize image-to-image translation techniques to narrow the gap between modalities distributions. Typically, these methods employ classical pixel manipulation techniques or deep neural networks to generate intermediate representations, which are then fed into a detector trained on the source modality. However, transitioning from IR to RGB has proven challenging due to generating color information while filtering out non-meaningful data associated with diverse heat sources. This challenge is particularly pronounced when the target category is also a heat-emitting source, such as a person.

In this work, we argue that achieving a robust intermediate representation for a given task needs guiding the image-to-image translation using a task-specific loss function. Here, we introduce HalluciDet, a novel approach for image translation focusing on detection tasks. Inspired by the learning using privileged information (LUPI) paradigm~\cite{vapnik2009new}, we utilize a robust people detection network previously trained on an RGB dataset to guide our translation process from IR to RGB. Our translation approach relies on an annotated IR dataset and an RGB detector to identify the appropriate representation space. The ultimate goal is to find a translation model, hereafter referred to as the Hallucination network, capable of translating IR images into meaningful representation to achieve accurate detections with an RGB detector. 

\noindent \textbf{Our main contributions can be summarized as follows:} \\

\noindent \textbf{(1)} We propose HalluciDet, a novel approach that leverages privileged information from pre-trained detectors in the RGB modality to guide end-to-end image-to-image translation for the IR modality.
\newline
\noindent \textbf{(2)} Given that our model focuses on the IR detection task, HalluciDet uses a straightforward yet powerful image translation network to reduce the domain gap between IR-RGB modalities, guided by the proposed hallucination loss function incorporating standard object detection terms.
\newline
\newline
\noindent \textbf{(3)} Through experiments conducted on two challenging IR-RGB datasets (LLVIP and FLIR ADAS), we compare HalluciDet against various image-to-image translation and traditional pixel manipulation methods. Our approach is seen to improve detection accuracy on the IR modality by incorporating privileged information from RGB.

\section{Related Work}
\label{sec:related_work}

\noindent \paragraph{Object detection.} Different from classification tasks, in which we want only to classify the object category, in object detection, additionally, the task is to know specific positions of the objects~\cite{zhang2021dive}. Deep learning object detection methods are categorized as two-stage and one-stage detectors. The two-stage detector extracts regions of interest or proposals for a second-stage classifier. Then, the second stage is responsible for classifying if there is an object in that region. One commonly used two-stage detector is the Faster R-CNN proposed by~\cite{ren2015faster}. It is the first end-to-end DL object detector to reach real-time speed. The speedup was achieved by introducing the Region Proposal Network (RPN), a network responsible for the region proposals without impacting the computational performance compared with previous region proposals algorithms~\cite{ren2016faster}. The one-stage detectors mainly focus on end-to-end training and real-time inference speed of the object detectors. In this scenario, the object detector has a single neural network to extract the features for the regression of the bounding box and give the class probabilities without an auxiliary network for the region proposals. Recently, there are detectors that were developed to remove the requirement of defining anchor boxes during training. For instance, the Fully Convolutional One-Stage Object Detection (FCOS) is one of these models that, due to its nature, reduces all complicated computation related to anchor boxes, which can lead to an increase in inference time.

\noindent \paragraph{Learning using Privileged Information (LUPI).} In human learning, the role of a teacher is crucial, guiding the students with additional information, such as explanations, comparisons, and so on~\cite{vapnik2009new}. In the LUPI setting, during the training, we have additional information provided by a teacher to help the learning procedure. Since the additional information is available at the training stage but not during the test time, we call it privileged information~\cite{vapnik2009new}. Recently,~\cite{lambert2018deep} proposed the usage of privileged information to guide the variance of a Gaussian dropout. In a classification scenario, additional localization information is used, and its results show that it improves the generalization, requiring fewer samples for the learning process~\cite{lambert2018deep}. \cite{motiian2016information} designed a large-margin classifier using information bottleneck learning with privileged information for visual recognition tasks. In the object detection problem,~\cite{hoffman2016learning} was the first work to present a modality hallucination framework, which incorporates the training RGB and Depth images, and during test time, RGB images are processed through the multi-modal framework to improve the performance of the detection. The modality hallucination network is responsible for mimicking depth mid-level features using RGB as input during the test phase.~\cite{liu2021depth} used depth as privileged information for object detection with a Depth-Enhanced Deformable Convolution Network. 
In this work, we use the privileged information coming from a pre-trained RGB detector to improve the performance of the infrared detection. In practice, instead of destroying the information of the RGB detector by fine-tuning, we use the RGB detector as a guide for translating the IR input image into a new representation, which can help the RGB detector boost performance by enhancing the objects of interest.

\noindent \paragraph{Image Translation.} The objective of image translation is to learn a mapping between two given domains such that images from the source domain can be translated to the target domain. In other words, the aim is to find a function $h_{\vartheta}: \mathcal{X}_{\text{s}} \rightarrow \mathcal{X}_{\text{t}}$ such that the distribution of images $h_{\vartheta}(\mathcal{X}_{\text{s}})$ in the translated domain is close to the distribution of images $\mathcal{X}_{\text{t}}$ in the target domain. Early methods rely on autoencoders (AEs)~\cite{hinton1993autoencoders} and generative adversarial networks (GANs)~\cite{goodfellow2014generative} to learn cross-domain mapping. Unsupervised AE methods aim to learn a representation of the data by reconstructing the input data. GANs are a type of generative model that can learn to generate new data that is similar to the training data. More recently, diffusion models have gained popularity. They are capable of generating high-quality images but lack some properties for domain translation, like on CycleGANs. For improving models such as CycleGAN, techniques such as Contrastive Unpaired Translation (CUT)~\cite{park2020contrastive} and FastCUT~\cite{park2020contrastive} were developed. CUT is an image translation model based on maximizing mutual information of patches, which is faster than previous methods while providing results as good as others. On RGB/IR modalities, the InfraGAN~\cite{ozkanouglu2022infragan} proposes an image-level adaptation using a model based on GANs, but for RGB to IR adaptation, with a focus on the quality of the generated images, thus optimizing image quality losses. Additionally,  using image translation for object detection on RGB/IR using pre-train models, Herrmann et al.~\cite{herrmann2018cnn} used RGB object detectors without changing their parameters. The IR images are adapted to the RGB images using traditional computer vision pre-processing at the image level before applying it as input to the RGB object detector.

None of these methods provides an end-to-end way to directly train the image translation methods for detection applications. Furthermore, traditionally, they require more than one kind of data set composed of the original domain and the target domain. For instance, CycleGAN is based on adversarial loss, and U-net is based on reconstruction loss. Thus, if we have access to the already trained detector on the original domain, this knowledge can possibly be used during the learning of the translation network.

\section{Proposed Method} 
\label{sec:method}

\begin{figure*}[t!]
	\centering
	\includegraphics[width=0.87\linewidth]{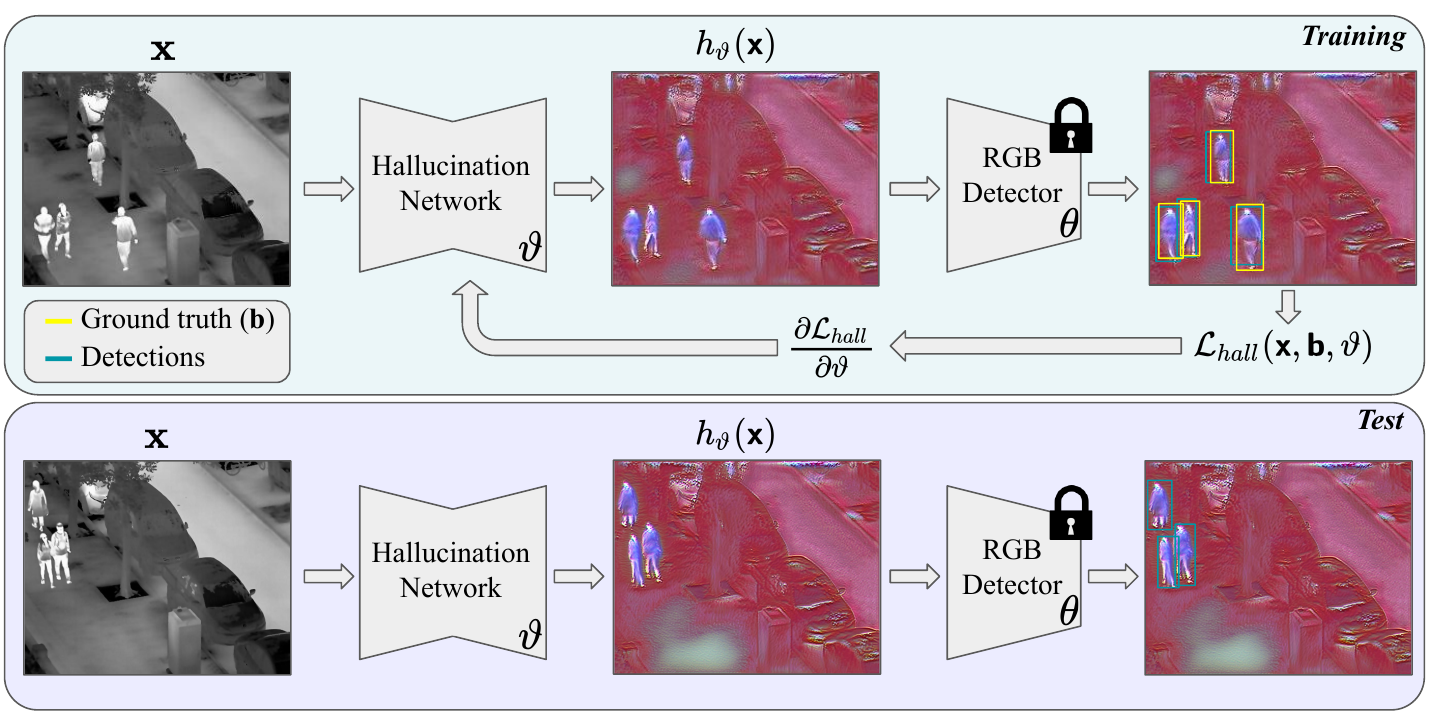}
\caption{HalluciDet leverages privileged information for modality hallucination with pre-trained detectors. During training, the hallucination network learns how to use the privileged information encoded by the RGB detector to translate the IR image into a new hallucination modality representation. Then, during inference, the model provides better IR detection using the translated modality.}
        \label{fig:hallucidet}
\end{figure*}

\noindent \paragraph{Preliminary definitions.} 
Let $\mathbf{x}_{\text{i}}$ be a given image with spatial resolution $W\times H$ and $C$ channels. An object detector aims to output a set of $N_{\text{reg}}$ object proposals, each represented as a bounding box $\textbf{b}_{\text{i,j}} = (c, d, e, w, h)$, where $(d,e)$ is the location of the top-left pixel of the bounding box for the $j$-th object, and $w$ and $h$ are the width and height of the object, respectively. Additionally, a classification label $c\in\{1, 2, ..., N_{\text{cls}}\}$ is assigned to each object of interest representing the region's class. In terms of optimization, such a task aims to maximize the detection accuracy, which typically is approximated through the average precision (AP) metric over all classes. Then, to train a detector, formally defined as the mapping $f_\theta\colon \mathbf{x}_{\text{i}} \rightarrow \hat{\mathbf{b}}_\text{i}$, a differentiable surrogate for AP metric is used, also known as the detection loss function, $\mathcal{L}_{\text{det}}(\textbf{b},\mathbf{x}; \theta)$.

The detection loss can be divided into two terms. The first one is the classification loss $\mathcal{L}_{\text{cls}}$($\hat{y}_{\text{c}}$,$y_{\text{c}}$) responsible for learning the class label $c$. In this work, we use the cross-entropy loss function to assess the matching of bounding boxes categories $\displaystyle\mathcal{L}_{\text{ce}}(\hat{y}_{\text{c}},y_\text{c}) = -\frac{1}{N_{\text{cls}}} \sum_{j=1}^{N_{\text{cls}}} y_{\text{c}_\text{j}}\log(p_{\text{j}})$, where $N_{\text{cls}}$ is the total number of classes, and $y_{\text{c}_\text{j}}$ is the class indicator function, i.e., $y_{\text{c}_\text{j}}=1$ if $c_\text{j}$ is the true class of the object, or $y_{\text{c}_\text{j}}=0$ otherwise. The probability provided by the detector for each category $j$ is $p_{\text{j}}$. To ensure the right positioning of the object, a second regression term $\mathcal{L}_{\text{reg}}$($\hat{\mathbf{y}}_{\text{b}}$,$\mathbf{y}_{\text{b}}$) is used, being the $\mathcal{L}_{\text{L1}}(\hat {\textbf{y}}_{\text{b}_\text{i}},\textbf{y}_{\text{b}_\text{i}}) = \sum_{i=1}^{N_{\text{reg}}}\left | \textbf{y}_{\text{b}_\text{i}} - \hat{\textbf{y}}_{\text{b}_\text{i}} \right | $ and $ \mathcal{L}_{\text{L2}}(\hat{\textbf{y}}_{\text{b}_\text{i}},\textbf{y}_{\text{b}_\text{i}}) = \sum_{i=1}^{N_{\text{reg}}}(\textbf{y}_{\text{b}_\text{i}} - \hat{\textbf{y}}_{\text{b}_\text{i}})^{2}$ losses the most commonly employed in the literature. Here $N_{\text{reg}}$ is the number of bounding boxes on the image $\textbf{x}_{\text{i}}$. Then, the final detection loss function can be defined in general terms as:  %
\begin{equation}
    \begin{split}
        \mathcal{L}_{\text{det}}(\mathbf{x},\textbf{b}; \theta) & = \mathcal{L}_{\text{cls}}(f_{\theta}(\mathbf{x}), c) \\
        & + \lambda \cdot \mathcal{L}_{\text{reg}}(f_{\theta}(\mathbf{x}), \textbf{b}),
    \end{split} 
    \label{eq:detection_loss}
\end{equation}
where $\lambda$ is a hyperparameter that controls the balance between the two terms, and $\theta$ is a vector containing the detector learnable parameters. The detectors used in this work use this general objective during their optimization process. However, they adapt each term to their specific architecture.

\noindent \paragraph{HalluciDet.} Our goal is to generate a representation from an IR image that a given RGB detector can effectively process.
Let $\mathcal{X}\subset\mathbb{R}^{W\times H}$ be the set of IR data containing $N$ images. During the learning phase, a training dataset $S=\{(\mathbf{x}_\text{i}, \mathbf{b}_\text{i})\}$ is given such that $\mathbf{x}_\text{i} \in \mathcal{X}$ is an IR image and $\mathbf{b}_\text{i}$ is a set of bounding boxes as defined in the previous section. In addition, an RGB detector $f_{\theta}$ is also available. Then, a representation mapping is here defined as $h_{\vartheta}\colon\mathcal{X}\to \mathcal{R}$, where $\mathcal{R}$ is the representation space and $\vartheta$ are the learnable parameters of the translation model. Such a representation space, $\mathcal{R}\subset \mathbb{R}^{W\times H \times 3}$, is conditioned to the subset of plausible RGB images that are sufficient to obtain a proper response from the RGB detector $f_{\theta}$. To find such a mapping we solve the optimization problem $\vartheta^* = \arg \min_{\vartheta} \mathcal{L}_{\text{hall}}(\mathbf{x},\textbf{b}; \vartheta)$ which implicitly uses the composition $(h_{\vartheta}\circ f_{\theta})(\mathbf{x})$ to guide the intermediate representation.

Our proposed model, HalluciDet, comprises two modules: a hallucination network responsible for the image-to-image harmonization and a detector. The Hallucination network is based on U-net~\cite{ronneberger2015u}, but modified with attention blocks which are more robust for image translation tasks~\cite{li2018pyramid, fan2020ma}. For training the HalluciDet, we train the hallucination module and condition it with the detection loss, which is the only supervision necessary for guiding the hallucination training with respect to the privileged information of the pre-trained RGB detector. This phase is responsible for translating the hallucinated image to a new representation close to the RGB modality. Please note that this strategy helps the final model to perform well on the IR modality without changing the knowledge from the detector. Under this framework, the RGB detection performance remains the same since the detector's parameters $\theta$ are not updated during the adaptation learning. On the other hand, detections over IR images are obtained by adapting the input using the Hallucination network, followed by the evaluation over the RGB detector. As a side advantage, our model allows evaluating both modalities by providing the appropriate modality identifier during the forward pass, i.e., RGB or IR. Figure \ref{fig:hallucidet} depicts the training and evaluation process of an IR image using privileged information from the RGB detector. 

The detector $f_{\theta}$ layers are frozen, thus preserving the prior knowledge, but the weights $\vartheta$ of the hallucination network $h_{\vartheta}$ are updated during the backward pass. The input minibatch is created with images from $\mathcal{X}$ set, leading to the hallucinated minibatch, which is then evaluated on $f_{\theta}$ to obtain the associated detections. To find the appropriate representation space, the hallucination loss $\mathcal{L}_{\text{hall}}(\textbf{x}, \textbf{b}, \vartheta)$ drives the optimization by updating only the hallucination network parameters. The representation space $\mathcal{R}$ is guided by $\mathcal{L}_{\text{hall}}$ to be closer enough to the RGB modality, which allows the detector to make successful predictions. As the representation is being learned with feedback from the frozen detector, it extracts the previous knowledge so that this new intermediate representation is tuned for the final detection task. The proposed hallucination loss shares some similarities with the aforementioned detection loss but with the distinction of only updating the modality adaptation parameters:
\begin{equation} 
\label{eq:hallucidet_first_step}
\begin{split}
\mathcal{L}_{\text{hall}}(\textbf{x},\textbf{b}, \vartheta) & = \mathcal{L}_{\text{cls}}(f_{\theta}(h_{\vartheta}(\textbf{x})), c) \\
 & + \lambda \cdot \mathcal{L}_{\text{reg}}(f_{\theta}(h_{\vartheta}(\textbf{x})), \textbf{b})
\end{split}
\end{equation}
Equation~\ref{eq:hallucidet_first_step} is optimized w.r.t $ \vartheta$. We added the hyperparameter $\lambda$ to weigh the contribution of each term and for numerical stability purposes.

\begin{table*}[h]
    \centering
    \begin{tabular}{@{}lcccccc@{}}

        \toprule
        
        \multirow{2}{*}[-1em]{\textbf{Image-to-image translation}} & \multirow{2}{*}[-1em]{\textbf{Learning strategy}}  &  \multicolumn{2}{c}{$\qquad\qquad\qquad$\textbf{AP@50$\uparrow$}} \\
        \cmidrule(lr){3-5}
        \addlinespace[5pt]
        
        {} & {}& \multicolumn{3}{c}{\multirow{2}{*}[1em]{\textbf{Test Set (Dataset: LLVIP)}}} \\
        
        {} & {} & FCOS  & RetinaNet & Faster R-CNN \\
                
        \midrule

        Blur~\cite{herrmann2018cnn} & - & 42.59 ± 4.17 & 47.06 ± 1.99 & 63.05 ± 1.96 \\

        Histogram Equalization~\cite{herrmann2018cnn} & -  & 33.10 ± 4.64	& 36.45 ± 2.02 & 51.47 ± 4.03 \\
        
        Histogram Stretching~\cite{herrmann2018cnn} & -     & 38.55 ± 4.25	& 41.97 ± 1.39 & 57.69 ± 2.78 \\
        
        Invert~\cite{herrmann2018cnn} & -  & 53.62 ± 2.07	& 55.43 ± 2.03 & 71.83 ± 3.04 \\

        Invert + Equalization~\cite{herrmann2018cnn} & -    & 50.03 ± 2.44	& 52.57 ± 1.50 & 68.69 ± 2.73 \\

        Invert + Equalization + Blur~\cite{herrmann2018cnn} & -   & 50.58 ± 2.41	& 52.62 ± 1.36 & 68.91 ± 2.74 \\
        
        Invert + Stretching~\cite{herrmann2018cnn} & - & 51.48 ± 2.17	& 52.87 ± 1.80 & 69.34 ± 3.07 \\

        Invert + Stretching + Blur ~\cite{herrmann2018cnn} & -   & 51.54 ± 1.92	& 52.96 ± 1.80 & 69.59 ± 2.90 \\
                
        Parallel Combination~\cite{herrmann2018cnn} & -  & 50.18 ± 2.25	& 52.52 ± 1.39 & 68.14 ± 2.98 \\
        
        \midrule
        
        U-Net~\cite{ronneberger2015u} & Reconstruction  & 42.94 ± 4.14 & 47.35 ± 1.92 & 63.23 ± 2.03 \\

        CycleGAN~\cite{zhu2017unpaired}& Adversarial & 22.76 ± 1.94 & 27.04 ± 4.23 & 38.92 ± 5.09 \\

        CUT~\cite{park2020cut}& Contrastive learning & 19.16 ± 2.10 & 21.61 ± 2.09 & 35.17 ± 0.32 \\
        
        FastCUT~\cite{park2020cut}& Contrastive learning & 46.87 ± 2.28 & 52.39 ± 2.31 & 67.73 ± 2.14 \\

        \rowcolor[HTML]{EFEFEF} 
        HalluciDet (ours)& Detection & \textbf{63.28 ± 3.49} & \textbf{56.48 ± 3.39} & \textbf{88.34 ± 1.50} \\
        
        \bottomrule
    \end{tabular}

\caption{Performance comparison of models on IR images using LLVIP dataset~\cite{jia2021llvip}. The table showcases the impact of different approaches, including pixel manipulation techniques, U-Net, CycleGAN, CUT, FastCUT, and HalluciDet. The detectors were trained with RGB data and evaluated on IR. To make a fair comparison with our models, we decided to start with models that do not have strong data augmentation that could benefit one modality over the other.}

\label{tab:5}
\end{table*}

\section{Experimental results and analysis}
\label{sec:experiments}

\noindent \paragraph{Experimental Methodology.}
Hallucidet is evaluated on two different popular IR/RGB datasets, the LLVIP~\cite{jia2021llvip}, and FLIR ADAS~\cite{flir2021free}. The LLVIP dataset is composed of $30,976$ images, in which $24,050$ ($12,025$ IR and $12,025$ RGB paired images) are used for training and $6,926$ for testing ($3,463$ IR and $3,463$ RGB paired images). For the FLIR, we used the sanitized and aligned paired sets provided by Zhang et al.~\cite{zhang2020multispectral}, which have $10,284$ images, being $8,258$ for training ($4,129$ IRs and $4,129$ RGBs) and $2,026$ ($1,013$ IRs and $1,013$ RGBs) for test. We chose to utilize these paired IR/RGB datasets to ensure a fair comparison with other image-to-image translation techniques that employ reconstruction losses. In our experiments, we use 80\% of the training set for training and the rest for validation. All results reported are on the test set. As for the FLIR dataset, we only used the person category. Initially, we have the RGB detector trained on the datasets using $5$ different seeds. It's worth noting that this model starts with pre-trained weights from COCO~\cite{lin2014microsoft}. Then with the RGB model trained, we use the model to perform the Hallucidet training. We tried ResNet$_{50}$ as the backbone for the detectors and ResNet$_{34}$ as the backbone for the Hallucination network. To ensure fairness we trained the detectors under the same conditions, i.e., data order, augmentations, etc. All the code is available at GitHub~\footnote{\url{https://github.com/heitorrapela/HalluciDet}.} for the reproducibility of the experiments. To develop the code, we used Torchvision models for the detectors and PyTorch Segmentation Models~\cite{Iakubovskii:2019} for the U-Net architecture of the hallucination network. Additionally, we trained with PyTorch Lightning~\cite{Falcon_PyTorch_Lightning_2019} training framework, evaluated the AP with TorchMetrics~\cite{detlefsen2022torchmetrics}, and logged all experiments with WandB~\cite{wandb} logging tool.

\noindent \paragraph{Main Comparative Results.} In~\tref{tab:5}, we investigate how our model behaved in comparison with standard image-to-image approaches and classical computer vision approaches that are normally used to reduce the distribution gap between IR and RGB. Furthermore, we highlight the impact of using the proposed $\mathcal{L}_{\text{hall}}$ loss to guide the representation. This is accomplished by comparing our approach with a U-Net that shares the same backbone as ours but employs a standard $\mathcal{L}_{\text{L1}}$ reconstruction loss. To guarantee comparability, we reproduce the experimental setting of~\cite{herrmann2018cnn} on our pipeline. We included basic pre-processing techniques that were shown to enhance IR performance on RGB models by Hermann et al.~\cite{herrmann2018cnn}. These techniques include a combination of blurring, histogram equalization, stretching, and inverting pixels. 
Furthermore, we included CycleGAN, which is a more powerful generative model compared with UNet. It is important to mention that training the CycleGAN is computationally more demanding than the Hallucidet. Additionally, due to the adversarial nature of the method, it does not ensure reliable convergence for the subsequent detection task. The CycleGAN was diverging with the same hyperparameters as~\cite{jia2021llvip} on the test set, so we tuned the hyperparameters and trained until the images became good qualitatively. Because CycleGAN introduces significant noise to the images as a result of its adversarial training, the detector's performance has notably decreased. This is particularly evident due to the increase in false positives. Given that our final goal is object detection, we selected FCOS, RetinaNet, and Faster R-CNN, each representing distinct categories within the universe of detection networks. We can see that straightforward approaches like inverting pixels for the IR and expanding it to three channels significantly enhance the initial performance of IR inputs on RGB detectors. As indicated in the table, our results demonstrate a significant improvement over previous image-to-image translation techniques in terms of detection performance. The most significant enhancement was observed in Faster R-CNN, where our proposal exhibited a remarkable 17\% improvement compared to pixel inversion.

\begin{figure*}[h]
\centering

    \begin{tabular}{c}
    
    \toprule

    \includegraphics[width=1.0\textwidth]{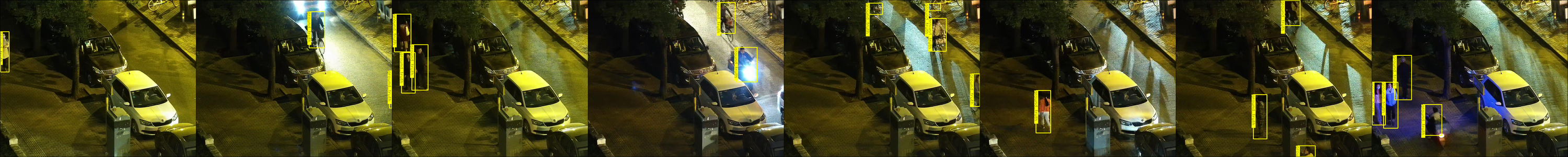} \\
    a) RGB - Ground Truth annotations.\\

    \includegraphics[width=1.0\textwidth]{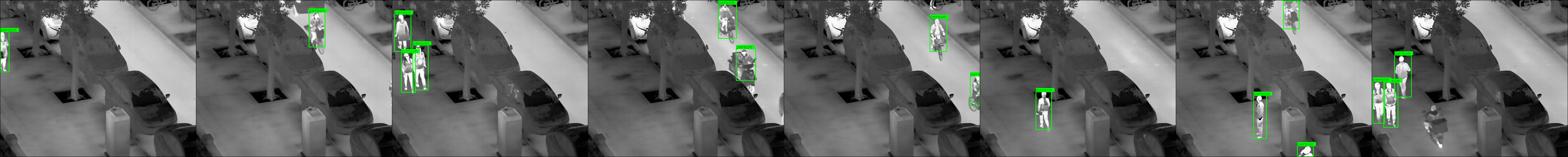} \\
    b) IR (Faster R-CNN) - Detections of the Fine-tuned model on the IR images. \\

    \includegraphics[width=1.0\textwidth]{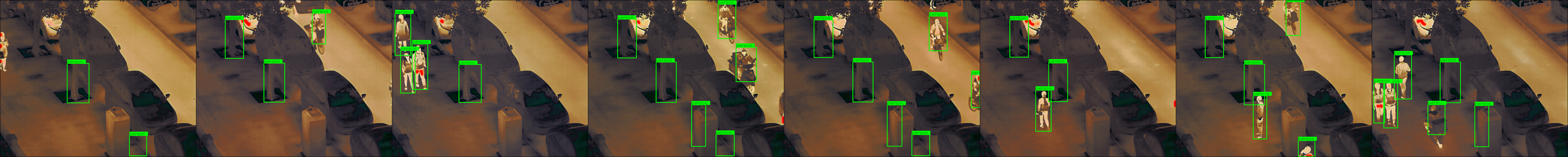} \\
    c) FastCUT (Faster R-CNN) - Detections of the RGB model on the transformed images. \\

    \includegraphics[width=1.0\textwidth]{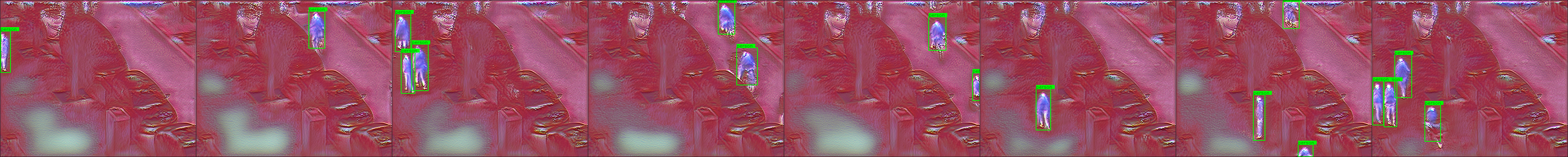} \\
    d) HalluciDet (Faster R-CNN) - Detections of the RGB model on the transformed images. \\

    \includegraphics[width=1.0\textwidth]{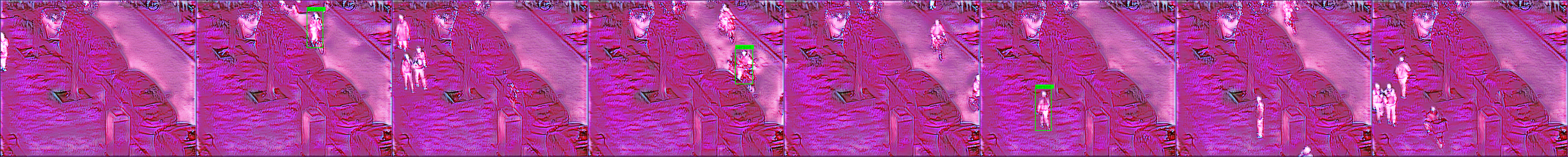} \\
    e) HalluciDet (FCOS) - Detections of the RGB model on the transformed images. \\
    
    \includegraphics[width=1.0\textwidth]{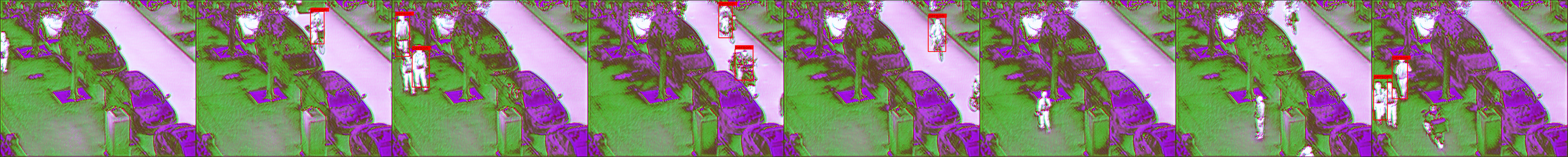} \\
    f) HalluciDet (RetinaNet) - Detections of the RGB model on the transformed images. \\

    \bottomrule

    \end{tabular}
\caption{Illustration of a sequence of $8$ images of LLVIP dataset. The first row is the RGB modality, then the IR modality, followed by FastCUT and different representations created by HalluciDet over various detectors.}

\label{fig:04.04_qualitative_experiments}

\end{figure*}

\noindent \paragraph{Hallucidet Visual Output.} In~\fref{fig:04.04_qualitative_experiments}, we present a Hallucination image and compare it with both RGB and IR. The Hallucination emphasizes the person while smoothing the background, helping the detector to distinguish the regions of interest. In contrast to RGB, our method allows for easy person detection even in low-light conditions. However, IR images may introduce additional non-person-related information that could bias the detector. A visual comparison with FastCUT is also provided, revealing a correlation between the method's low performance and the high number of False Positives detected. It is important to note that while we show the Hallucination for representation demonstration, our main goal is on detection metrics. In the figure, the ground truth bounding box annotations are shown in yellow on the RGB images. The corresponding detections obtained from the IR data are presented in the following lines. It is important to note that we display the predicted detections on top of the intermediate representation for convenience. However, the actual inputs for HalluciDet approaches and FastCUT are IR images. A significant number of False Positives can be observed for FastCUT, while HalluciDet (FCOS) and HalluciDet (RetinaNet) exhibit a high number of False Negatives. The most accurate detection results are achieved with HalluciDet (Faster R-CNN), which demonstrates superior performance to the IR fine-tuned model in cases where the person's heat signature is not clearly evident, as seen in the last column. Additional figures can be found in the supplementary material.

\noindent \paragraph{Comparison with fine-tuning.} For this experiment, we performed an evaluation of both RGB and fine-tuned IR detectors that were trained on the LLVIP and FLIR datasets. All methods from Table~\ref{tab:2} were trained under the same experimental protocol using $3$ different seeds.

Similar to the previous experiment, we utilized a detector from each family of methods, namely FCOS, RetinaNet, and Faster R-CNN. The provided results include the mean and standard deviation of the AP on the test set. 
In this experiment, we compare three different approaches to adapt a model trained on RGB images to IR. As baseline we consider the case of No Adaptation, in which the model is used directly on IR images. Then, we consider the case in which a model is adapted to the IR data with normal fine-tuning, which is the most common way of adaptation when annotations are available. Finally, we train our HalluciDet to generate a new representation of the image for the RGB detector.

As seen in Table~\ref{tab:2}, in all cases, the fine-tuned IR model outperformed the RGB detector over the IR modality, as expected. In the tables, we also observe a significant improvement in the performance of HalluciDet compared to the performance achieved through fine-tuning for Faster R-CNN. This improvement aligns with the quality of the representation observed in Figure~\ref{fig:04.04_qualitative_experiments}, where confusing factors, such as car heat, have been removed from the image. A marginal improvement was observed with center point-based architectures like FCOS for the LLVIP dataset, although a higher difference in AP could be observed for the FLIR dataset. On the other hand, the results using RetinaNet didn't exhibit much consistency; the AP was significantly worse than that achieved through fine-tuning for the LLVIP dataset. Once again, this is consistent with the observed representation lacking the necessary discriminative information to detect people in the image.

\begin{table}[h]
    \centering
    \resizebox{1.0\columnwidth}{!}{%
    \begin{tabular}{@{}lcccc@{}}

        \toprule
        
        \multirow{2}{*}[-1em]{\textbf{Method}} & \multicolumn{3}{c}{\textbf{AP@50$\uparrow$}} \\
        \cmidrule(lr){2-4}
        \addlinespace[5pt]
        
        {}  & \multicolumn{3}{c}{\multirow{2}{*}[1em]{\textbf{Test Set IR (Dataset: LLVIP)}}} \\
        
        {} & No Adaptation  & Fine-tuning & HalluciDet\\
        
        \midrule

        FCOS & 47.12 ± 4.32 & 63.79 ± 0.48 & \textbf{64.85 ± 1.46}  \\

        \midrule
        
        RetinaNet  & 50.63 ± 3.22 & \textbf{76.26 ± 0.75} & 56.78 ± 3.85  \\
         
        \midrule
        
        Faster R-CNN   & 71.51 ± 1.16 & 84.94 ± 0.15 & \textbf{90.92 ± 0.20}  \\

        \midrule

        \addlinespace[5pt]
        
        {}  & \multicolumn{3}{c}{\multirow{2}{*}[1em]{\textbf{Test Set IR (Dataset: FLIR)}}} \\
        
        {} & No Adaptation  & Fine-tuning & HalluciDet\\
        
        \midrule

        FCOS & 38.52 ± 0.79 & 42.22 ± 1.04 & \textbf{49.18 ± 0.99}  \\
        
        \midrule
    
        RetinaNet  & 44.13 ± 2.01 & 47.87 ± 2.21 & \textbf{49.01 ± 4.08}  \\

        \midrule

        Faster R-CNN   & 55.85 ± 1.19 & 61.48 ± 1.55  & \textbf{70.90 ± 1.35} \\

        \bottomrule
    \end{tabular}
    }
\caption{AP performance for various models following distinct training approaches on two datasets of LLVIP \cite{jia2021llvip} (top half) and FLIR \cite{fa2018flir} (bottom half): starting from COCO pre-training and fine-tuning on the RGB data shown as (No Adaptation) and fine-tuning on the IR data shown as (Fine-tuning). In the case of HalluciDet, the trained RGB detector serves as the initial point, with the subsequent optimization of the Hallucination network using the IR data. The reported performance is exclusive to the person category.}
\label{tab:2}
\end{table}

\paragraph{Hallucidet with different backbones.} In Table~\ref{tab:llvip_diff_backbones}, we investigated various encoder backbones for the Hallucination network. The presented results include two MobileNet and two ResNets with different widths. Additional outcomes for alternative backbones are included in the supplementary material. In all cases, the model consistently improves upon the performance of the fine-tuned IR model. Notably, even in models with a reduced number of parameters, such as MobileNet$_{v2}$ with less than 7 million additional parameters, the gain remains consistent at nearly 5\%.

\begin{table}[h]
    \centering
    \resizebox{0.8\columnwidth}{!}{%
    
    \begin{tabular}{@{}clcc@{}}

        \toprule

        \multicolumn{2}{c}{\textbf{Method}}  & \textbf{Params.} &  \textbf{AP@50$\uparrow$} \\
        \midrule
        \addlinespace[5pt]
        \multicolumn{2}{c}{Faster R-CNN}  & 41.3 M & 84.83 \\
        \midrule 
        \multirow{2}{*}[-1em]{HalluciDet} & MobileNet$_{v3s}$ & + 3.1 M & 85.20 \\
        {} & MobileNet$_{v2}$ & + 6.6 M & 89.73 \\
        {} & ResNet$_{18}$ & + 14.3 M & 90.42 \\
        {} & ResNet$_{34}$ & + 24.4 M  & 90.65 \\
        
        \bottomrule
    \end{tabular}
    }
\caption{Comparison of the number of parameters for different Hallucination Network backbones vs. AP@50 on the LLVIP dataset with the Faster R-CNN detector.}
\label{tab:llvip_diff_backbones}
\end{table}

\begin{figure}[h]
\centering
    \resizebox{\columnwidth}{!}{%

    \begin{tabular}{c}
    
    \includegraphics[width=0.75\columnwidth]{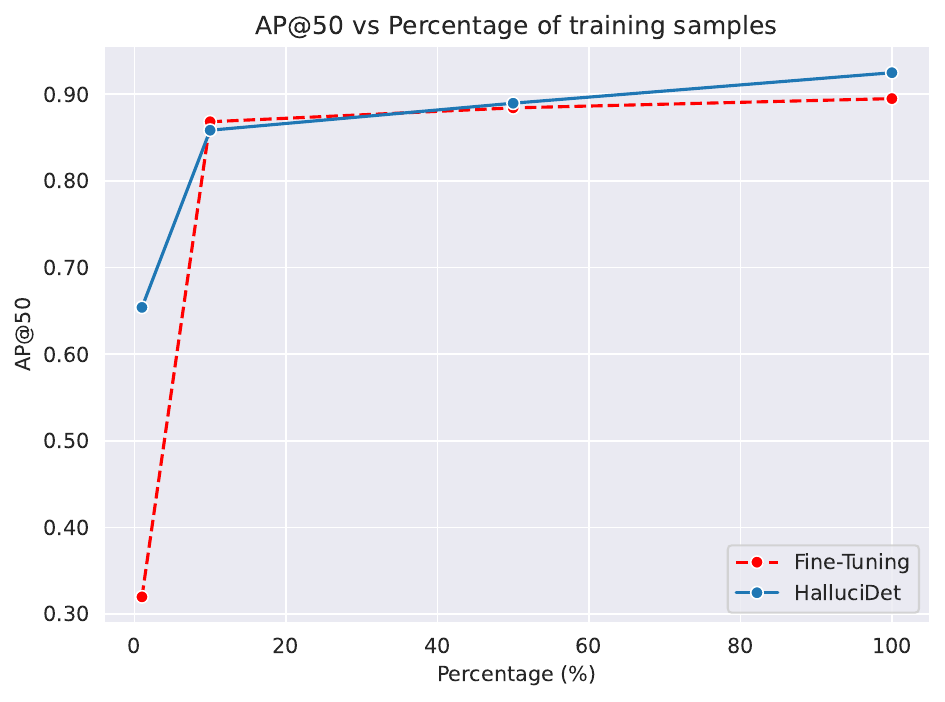} 

    \end{tabular}
    }
\caption{AP@50 vs. training samples percentages. The figure shows the AP@50 over the LLVIP test set using various amounts of training samples for the HalluciDet Faster R-CNN.}
\label{fig:testset_mAP_varying_training_samples}
\end{figure}

\begin{figure}[h]
\centering
    \resizebox{\columnwidth}{!}{%

    \begin{tabular}{c}
    
    \includegraphics[width=0.75\columnwidth]{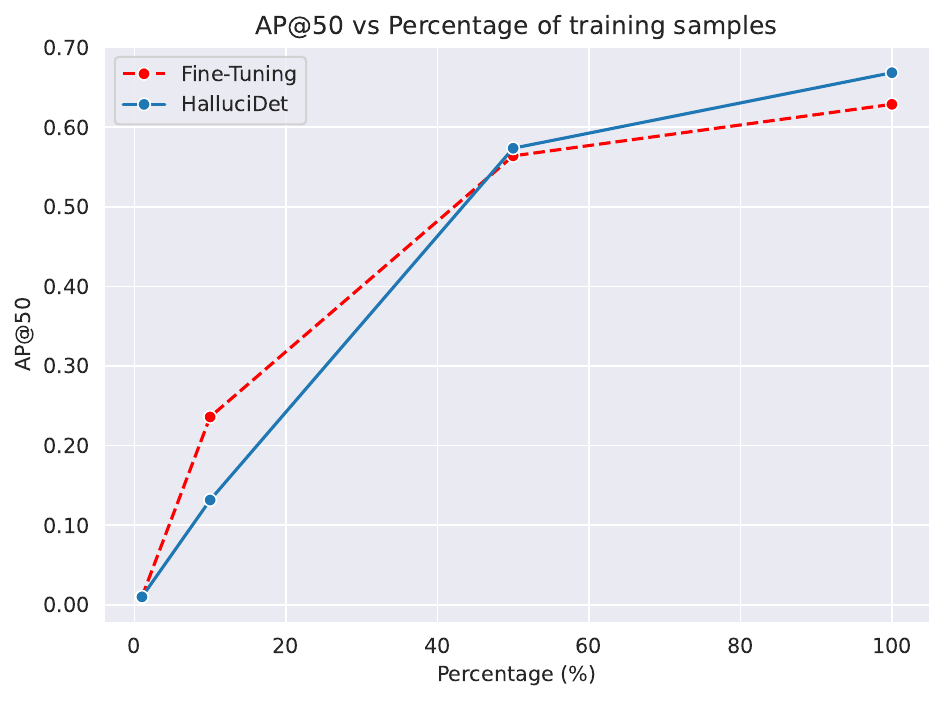} 

    \end{tabular}
    }
\caption{AP@50 vs. training samples percentages. The figure shows the AP@50 over the FLIR test set using various amounts of training samples for the HalluciDet Faster R-CNN. Notably, 70\% of the data was sufficient for HalluciDet to achieve comparable performance to the fine-tuned Faster R-CNN with the complete dataset.}
\label{fig:flir_testset_mAP_varying_training_samples}
\end{figure}

\noindent \paragraph{Hallucidet with a different number of training samples.} For the LLVIP dataset, in Figure~\ref{fig:testset_mAP_varying_training_samples}, we explored various quantities of training samples for our method, ranging from 1\% to 100\%. Notably, only 30\% of the data was sufficient for HalluciDet to achieve comparable performance to the fine-tuned Faster R-CNN with the
complete dataset. For the FLIR dataset, in Figure~\ref{fig:flir_testset_mAP_varying_training_samples}, the trend to reduce the number of training samples and improve over the fine-tuning is still true, but in this case, around 70\% of the training samples. The different characteristics related to the exact number of training samples with respect to the dataset are due to the number of different environment changes on the datasets. For the LLVIP, we do not have a big shift in the images because the cameras are fixed in a surveillance context. In the case of FLIR, the variance of the images is higher due to the different capture settings; with the focus on autonomous driving, the camera moves inside a car, which changes the background consistency and introduces more variance to the dataset.

\section{Conclusion}
\label{sec:conclusion}

In this work, we provided a framework that uses privileged information of an RGB detector to perform the image-to-image translation from IR. The approach involves utilizing a Hallucination network to generate intermediate representations from IR data, which are then directly input into an RGB detector. An appropriate loss function was also proposed to lead the representation into a space that allows for the enhancement of the target category's importance. 

In our experiments, we demonstrate that hallucination networks can be helpful for modality adaptation by obtaining an intermediate representation that effectively supports accurate responses in the object detection task. The proposed approach showed particular effectiveness for the two-stage detector Faster R-CNN, resulting in a reduction of non-person-related information. This reduction in background clutter had a positive effect on minimizing the number of False Positives, surpassing the performance of standard fine-tuning on IR data. The comparison with methods from the literature for image-to-image translation highlighted the significance of guiding the representation to achieve successful detections. Our Hallucidet demonstrated a significant performance improvement compared to the other methods. Finally, the proposed framework offers the additional advantage of maintaining performance in the RGB task, which is beneficial for applications requiring accurate responses in both modalities.

\textbf{Acknowledgments}: This work was supported by Distech Controls Inc., the Natural Sciences and Engineering Research Council of Canada, the Digital Research Alliance of Canada, and MITACS.

\section{Supplementary Material: \\ 
HalluciDet: Hallucinating RGB Modality for Person Detection Through Privileged Information}  %

\maketitle
\thispagestyle{empty}
\appendix

In this supplementary material, we provide additional information to reproduce our work. The source code\footnote{\url{https://github.com/heitorrapela/HalluciDet}.} is publicly provided. Here, we provide ablation on the hyperparameter of the HalluciDet loss, qualitative examples of the obtained detections, and additional results.

\section{Ablation of hyperparameters $\lambda$ for HalluciDet}

In this section, we show the sensitivity of HalluciDet to the hyperparameters during training. For these experiments, as we did not want to have the influence of data augmentation on the pipeline, we removed the data augmentations that could benefit the starting detector and also the HalluciDet. Thus, the results in the main manuscript are the results with the detector using transformations such as color jitter and horizontal flip, and the same transformations were used for training the HalluciDet and respective baselines. In this ablation, we focused on the balancing of $\lambda$, so for this case, we kept both detectors and HalluciDet without data augmentations.

The cost function of the hallucination network $\mathcal{L}_{hall}$ (\eref{eq:hall_loss}) contains three terms: regression loss, classification loss, and other losses. Here, the other loss terms are dependent on the detection method used, e.g., for Faster-RCNN $\mathcal{L}_* = \mathcal{L}_{rpn} + \mathcal{L}_{obj}$, where the regression loss $\mathcal{L}_{rpn}$ is applied to the region proposal network, and $\mathcal{L}_{obj}$ is the object/background classification loss.

\begin{equation} 
\label{eq:hall_loss}
\begin{split}
\mathcal{L}_{\text{hall}} = \lambda_{cls} \cdot \mathcal{L}_{\text{cls}} + \lambda_{reg} \cdot \mathcal{L}_{\text{reg}} + \lambda_{*} \cdot \mathcal{L}_{*}
\end{split}
\end{equation}

As shown  on~\tref{tab:1} of this supplementary materials, the HalluciDet ablation study was divided into different ways of balancing the regression and classification parts of the loss. In practice, it is better to use both components (regression and classification), but we recommend prioritizing the regression part for optimal balance.

\begin{table*}[ht]
    \centering
    \resizebox{\textwidth}{!}{%
    \begin{tabular}{ll|c}
    
        \toprule

        \multirow{2}{*}[-1em]{\textbf{Method}} & \multirow{2}{*}[-1em]{\hspace{0.5cm}\textbf{Ablation (Loss Weight)}} &  \multicolumn{1}{c}{\textbf{AP@0.5$\uparrow$}} \\
        \cmidrule(lr){3-3}
        \addlinespace[5pt]
        
        {} & {} & \multicolumn{1}{c}{\multirow{2}{*}[1em]{\textbf{Test Set (Dataset: LLVIP)}}} \\
        
        {} & {} & $\mathcal{X}^{IR}$ \\

        \midrule
        
        \multirow{5}{*}{HalluciDet (RetinaNet)} &  $\qquad\lambda_{cls}=0.0$, $\lambda_{reg}=1.0$  & 65.81 \\ 
        \cmidrule(lr){2-2}

        {} & $\qquad\lambda_{cls}=1.0$, $\lambda_{reg}=0.0$  & 60.77 \\

        \cmidrule(lr){2-2}
        
        {} & $\qquad\lambda_{cls}=0.01$, $\lambda_{reg}=0.1$  & 68.58 \\

        \cmidrule(lr){2-2}
        
        {} & $\qquad\lambda_{cls}=0.1$, $\lambda_{reg}=0.01$  & 60.03 \\

        \midrule

        \multirow{5}{*}{HalluciDet (FCOS)} &  $\qquad\lambda_{cls}=0.0$, $\lambda_{reg}=1.0$, $\lambda_{box_{cnt}}=1.0$  & 63.01 \\ 
        \cmidrule(lr){2-2}

        {} & $\qquad\lambda_{cls}=1.0$, $\lambda_{reg}=0.0$, $\lambda_{box_{cnt}}=0.0$  & 60.92 \\

        \cmidrule(lr){2-2}
        
        {} & $\qquad\lambda_{cls}=0.01$, $\lambda_{reg}=0.1$,  $\lambda_{box_{cnt}}=0.1$  & 65.02 \\

        \cmidrule(lr){2-2}
        
        {} & $\qquad\lambda_{cls}=0.1$, $\lambda_{reg}=0.01$, $\lambda_{box_{cnt}}=0.01$  & 64.59 \\

        \midrule

        \multirow{5}{*}{HalluciDet (Faster R-CNN)} &  $\qquad\lambda_{cls}=0.1$, $\lambda_{obj}=0.1$, $\lambda_{reg}=0.01$, $\lambda_{RPNbox_{reg}}=0.01$  & 85.35 \\ 
        \cmidrule(lr){2-2}

        {} & $\qquad\lambda_{cls}=0.01$, $\lambda_{obj}=0.01$, $\lambda_{reg}=0.1$, $\lambda_{RPNbox_{reg}}=0.1$  & 88.72 \\ 

        \cmidrule(lr){2-2}
        
        {} & $\qquad\lambda_{cls}=1.0$, $\lambda_{obj}=1.0$, $\lambda_{reg}=0.0$, $\lambda_{RPNbox_{reg}}=0.0$  & 83.97 \\ 

        \cmidrule(lr){2-2}
        
        {} & $\qquad\lambda_{cls}=0.0$, $\lambda_{obj}=0.0$, $\lambda_{reg}=1.0$, $\lambda_{RPNbox_{reg}}=1.0$  & 84.08 \\

        \bottomrule
    \end{tabular}
    }
\caption{Comparison between different weights on the losses terms. In this table, the models are started frozen from RGB, the same as reported in the paper. Then, the hallucination network is trained with different lambda values to see its impacts on the model's performance. Results over LLVIP test set.}
\label{tab:1}
\end{table*}

\section{Hallucidet and additional results on FLIR}

Similar to the main manuscript, we added additional ablations with respect to the FLIR dataset.

\noindent \paragraph{Hallucidet with a different encoder.} Similar to the main manuscript, we provided a study on the different backbones of the hallucination network encoder but focused on the FLIR dataset. The results show a similar trend, in which models with more capacity in terms of parameters can learn more robust representations for the test set distribution, thus increasing the AP@50.

\begin{table}[h]
    \centering
    \resizebox{0.8\columnwidth}{!}{%
    
    \begin{tabular}{@{}clcc@{}}

        \toprule

        \multicolumn{2}{c}{\textbf{Method}}  & \textbf{Params.} &  \textbf{AP@50$\uparrow$} \\
        \midrule
        \addlinespace[5pt]
        \multicolumn{2}{c}{Faster R-CNN}  & 41.3 M & 61.48 \\
        \midrule 
        \multirow{2}{*}[-1em]{HalluciDet} & MobileNet$_{v3s}$ & + 3.1 M & 53.62 \\
        {} & MobileNet$_{v2}$ & + 6.6 M & 67.74 \\
        {} & ResNet$_{18}$& + 14.3 M & 68.56 \\
        {} & ResNet$_{34}$ & + 24.4 M  & 71.58 \\
        
        \bottomrule
    \end{tabular}
    }
\caption{Comparison of the number of parameters for different Hallucination Network backbones vs. AP@50 on the FLIR dataset with the Faster R-CNN detector.}
\label{tab:x}
\end{table}

\section{Qualitative analysis of Hallucidet Detections}

\begin{figure*}[h]
\centering

    \begin{tabular}{c}
    
    \toprule

    \includegraphics[width=1.0\textwidth]{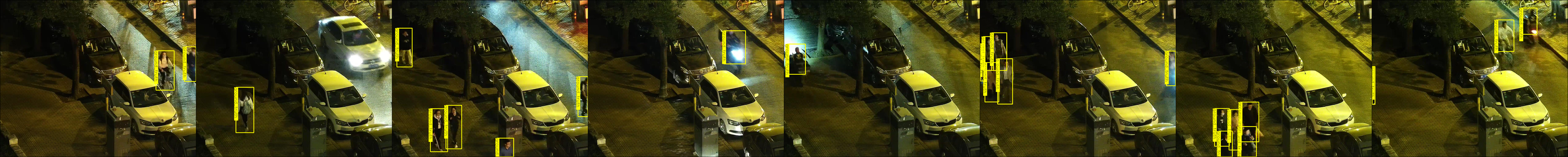} \\
    a) RGB - Ground Truth annotations.\\

    \includegraphics[width=1.0\textwidth]{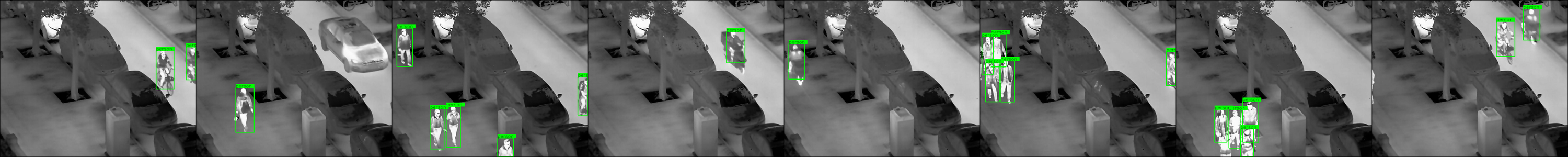} \\
    b) IR (Faster R-CNN) - Detections of the Fine-tuned model on the IR images. \\

    \includegraphics[width=1.0\textwidth]{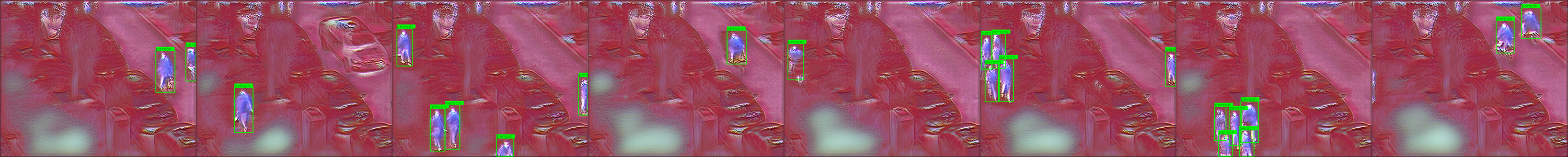} \\
    c) HalluciDet (Faster R-CNN) - Detections of the RGB model on the transformed images. \\

    \includegraphics[width=1.0\textwidth]{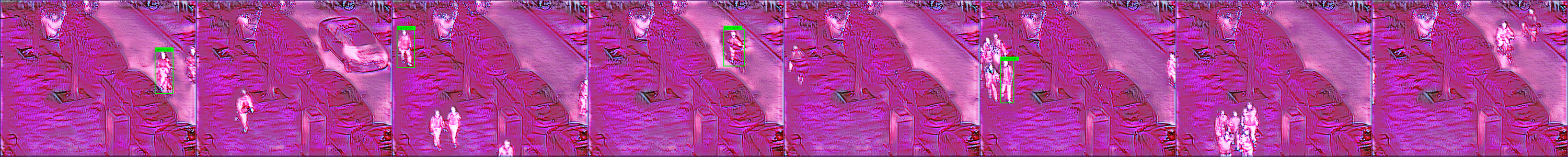} \\
    d) HalluciDet (FCOS) - Detections of the RGB model on the transformed images. \\
    
    \includegraphics[width=1.0\textwidth]{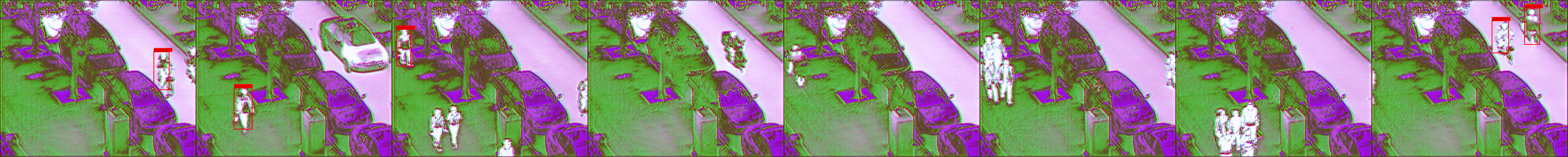} \\
    e) HalluciDet (RetinaNet) - Detections of the RGB model on the transformed images. \\

    \bottomrule

    \end{tabular}
\caption{Illustration of a sequence of $8$ images of LLVIP dataset. The first row is the RGB modality, then the IR modality, followed by different representations created by HalluciDet over various detectors.}

\label{fig:qe_llvip_01}

\end{figure*}

\begin{figure*}[h]
\centering

    \begin{tabular}{c}
    
    \toprule

    \includegraphics[width=1.0\textwidth]{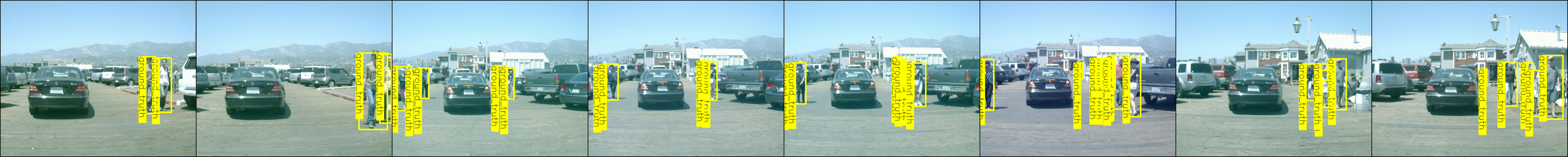} \\
    a) RGB - Ground Truth annotations.\\

    \includegraphics[width=1.0\textwidth]{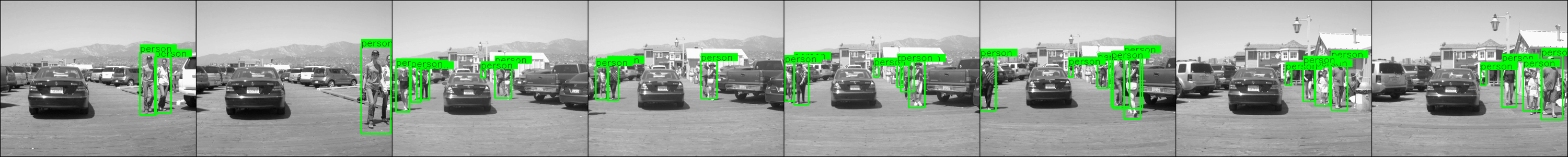} \\
    b) IR (Faster R-CNN) - Detections of the Fine-tuned model on the IR images. \\

    \includegraphics[width=1.0\textwidth]{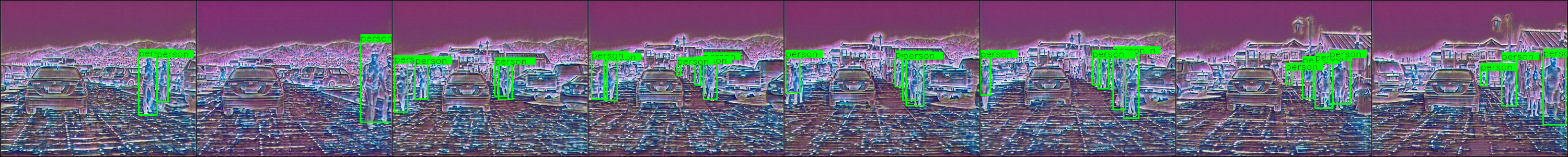} \\
    c) HalluciDet (Faster R-CNN) - Detections of the RGB model on the transformed images. \\

    \includegraphics[width=1.0\textwidth]{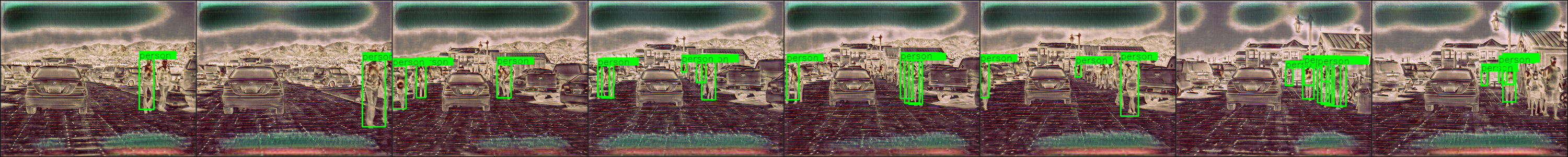} \\
    d) HalluciDet (FCOS) - Detections of the RGB model on the transformed images. \\
    
    \includegraphics[width=1.0\textwidth]{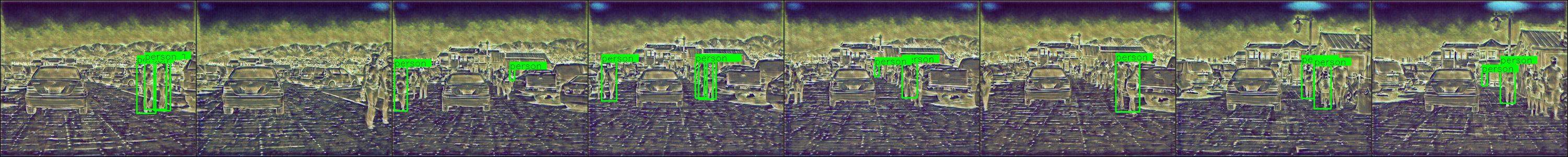} \\
    e) HalluciDet (RetinaNet) - Detections of the RGB model on the transformed images. \\

    \bottomrule

    \end{tabular}
\caption{Illustration of a sequence of $8$ images of FLIR dataset. The first row is the RGB modality, then the IR modality, followed by different representations created by HalluciDet over various detectors.}

\label{fig:qe_flir_01}

\end{figure*}

\begin{figure*}[h]
\centering

    \begin{tabular}{c}
    
    \toprule

    \includegraphics[width=1.0\textwidth]{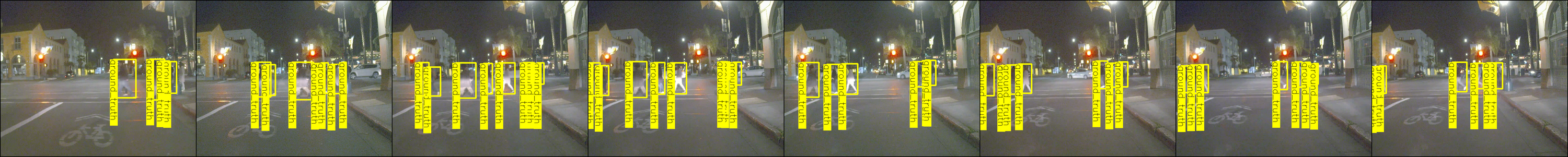} \\
    a) RGB - Ground Truth annotations.\\

    \includegraphics[width=1.0\textwidth]{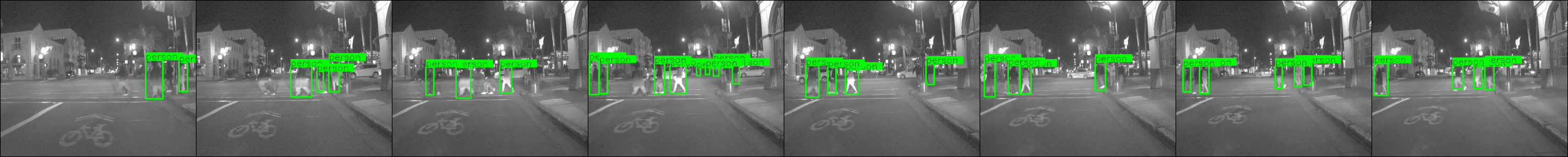} \\
    b) IR (Faster R-CNN) - Detections of the Fine-tuned model on the IR images. \\

    \includegraphics[width=1.0\textwidth]{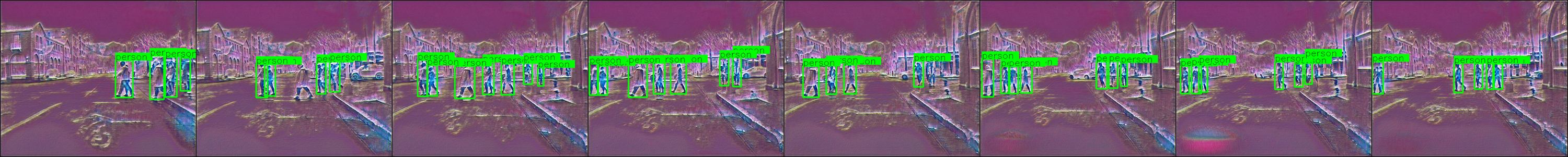} \\
    c) HalluciDet (Faster R-CNN) - Detections of the RGB model on the transformed images. \\

    \includegraphics[width=1.0\textwidth]{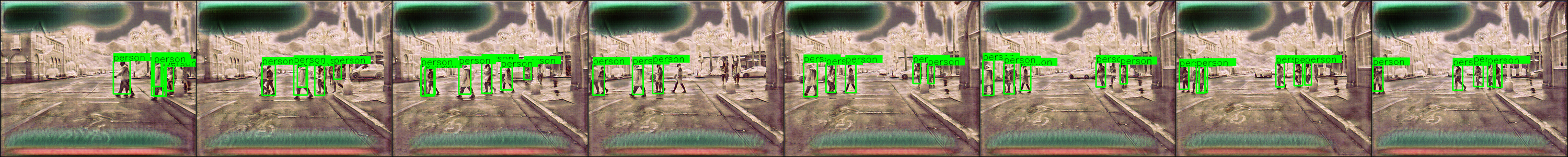} \\
    d) HalluciDet (FCOS) - Detections of the RGB model on the transformed images. \\
    
    \includegraphics[width=1.0\textwidth]{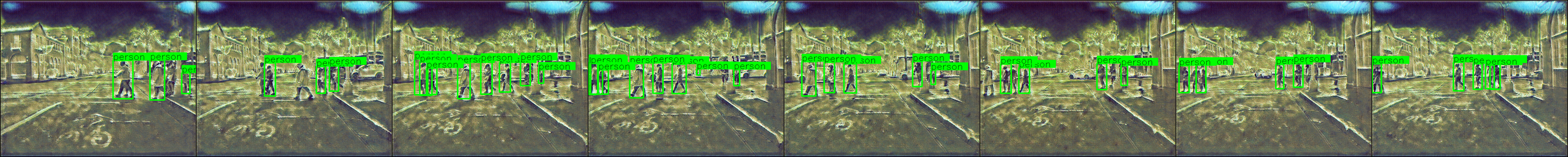} \\
    e) HalluciDet (RetinaNet) - Detections of the RGB model on the transformed images. \\

    \bottomrule

    \end{tabular}
\caption{Another sequence of $8$ images of FLIR dataset. The first row is the RGB modality, then the IR modality, followed by different representations created by HalluciDet over various detectors.}

\label{fig:qe_flir_02}

\end{figure*}

In this section, we provided an additional sequence of batch images, similar to the main manuscript. Here, we can find more than one batch of $8$ images for the LLVIP dataset (\fref{fig:qe_llvip_01}), and then two batches of $8$ images each for the FLIR dataset (\fref{fig:qe_flir_01}, \fref{fig:qe_flir_02}). Thus, the trend and explanations for detections remain the same as those described in the main manuscript.

\noindent \textbf{Processing time comparison}: In terms of trade between more parameters that can increase the speed for processing and the performance of the detection, we highlight some important discussion about it. For the models classified as nonlearning methods, there is no increase in the inference speed and the training part, but they have lower detection performance. For the models that are learning in the input space, such as image translation methods like CycleGAN or FastCUT, given the same backbone network, HalluciDet has faster training and equal inference time to the deep learning baselines, and we can improve the detection performance.

{\small
\bibliographystyle{ieee_fullname}
\bibliography{main}
}

\end{document}